\documentclass{article}

\PassOptionsToPackage{numbers, compress}{natbib}
\usepackage[preprint]{neurips_2026}
\usepackage[utf8]{inputenc}
\usepackage[T1]{fontenc}
\usepackage{hyperref}
\usepackage{url}
\usepackage{booktabs}
\usepackage{graphicx}
\usepackage{amsmath}
\usepackage{amsfonts}
\usepackage{nicefrac}
\usepackage{microtype}
\usepackage{xcolor}
\usepackage{soul}
\usepackage{multirow}
\usepackage{bm}
\usepackage{makecell}
\usepackage[dvipsnames]{xcolor}

\title{MAG-VLAQ: Multi-modal Aerial-Ground Query Aggregation for Cross-View Place Recognition}

\author{%
  Zhengyi~Xu\thanks{Equal Contribution.}$^{*1}$ Yuhang~Ming$^{*}$\thanks{Corresponding Author.}$^{\dagger1}$ Zhihao Zhan$^{*2}$ Hanyu Zhu$^{1}$ Javier Civera$^{3}$ Wanzeng Kong$^{1}$
  \\
  $^1$Hangzhou Dianzi University\quad
  $^2$TopXGun Robotics\quad
  $^3$University of Zaragoza\\
  \texttt{\{jasperxzy, yuhang.ming, kongwanzeng\}@hdu.edu.cn} \\
}

\begin{document}

\maketitle

\begin{abstract}
Multi-modal cross-view place recognition remains a fundamental challenge in computer vision and robotics due to the severe viewpoint, modality, and spatial-structure discrepancies between ground observations and aerial references.
To address this challenge, we present \textbf{MAG-VLAQ}, a foundation-model-enhanced query aggregation framework for multi-modal aerial-ground cross-view place recognition. 
Specifically, our approach leverages pre-trained foundation models to extract dense visual tokens from both ground and aerial images, as well as expressive geometric tokens from ground LiDAR observations. These heterogeneous tokens are then projected into a shared embedding space for cross-modal alignment and fusion. As our main contribution, we propose ODE-conditioned VLAQ, which 
tightly couples neural ordinary differential equations (ODE)-based RGB-LiDAR fusion with vectors of locally aggregated queries (VLAQ). In this design, the VLAQ query centers are dynamically adapted according to the fused multi-modal state.
This mechanism allows the final global descriptor to preserve globally learned retrieval prototypes while remaining responsive to scene-specific visual and geometric evidence, significantly improving aerial-ground matching. 
Extensive experiments on KITTI360-AG and nuScenes-AG validate the effectiveness of our proposed MAG-VLAQ. Notably, on KITTI360-AG, our MAG-VLAQ nearly doubles the state-of-the-art performance, achieving \textbf{61.1} Recall@1 in the satellite setting, compared with \textbf{34.5} from the closest competing approach.
\end{abstract}

\section{Introduction}
Place recognition is a fundamental capability for long-term robot localization~\citep{zaffar2023copr}, loop closure detection~\citep{maggio2025vggt}, autonomous navigation~\citep{Suomela_2024_Placenav}, and large-scale spatial understanding~\citep{lindenberger2025scaling}. 
Given a query observation, the goal is to retrieve database samples that correspond to the same physical location, \textit{i.e.} recognize the place, thereby formulating localization as a retrieval problem~\citep{lowry_visual_2016,Garg_2021_Survey,schubert2023visual,milford2025going}. 

Over the past decade, place recognition has witnessed substantial progress in conventional ground-view settings. 
A large body of work has improved recognition performance by designing stronger feature representations, ranging from convolutional backbones and local feature learning~\cite{arandjelovic_netvlad_2016, uy_pointnetvlad_2018, komorowski_minkloc_2021, zhou_lcpr_2024, ming_cgisnet_2022, ming_aegisnet_2024} to transformer-based and self-supervised visual representations~\citep{keetha_anyloc_2023, lu_cricavpr_2024, lu_selavprplusplus_2025, izquierdo_close_2024}. 
In parallel, another important line of research focuses on global descriptor aggregation, where local visual or geometric features are compressed into compact and discriminative retrieval descriptors. 
Representative designs include pooling-based aggregation~\cite{radenovic_gem_2019, ali-bey_mixvpr_2023}, clustering-based aggregation~\cite{sivic_bow_2009, ali-bey_boq_2024, izquierdo_salad_2024} , residual-based aggregation~\cite{jegou_vlad_2010, arandjelovic_netvlad_2016, zhu_dcvlaq_2026}, and geometry-based descriptors~\cite{lu_ring_2022, xu_ring++_2023}. 
Together, these advances have pushed place recognition performance to a high level on many standard benchmarks~\cite{olid_nordland_2018, torii_tokyo247_2018, warburg_msls_2020, torii_pitts_2013}. 

However, aerial-ground cross-view place recognition remains comparatively under-explored, where ground observations from onboard sensors must be matched against aerial references such as satellite images or road maps~\citep{hu_cvm-net_2018, samano_youarehere_2020}. 
This setting requires aligning ground-level visual and geometric cues with top-down scene representations under substantial viewpoint, scale, appearance, modality, and layout discrepancies.
A key practical challenge is that a monocular camera captures only a limited field of view, lacking the broad spatial context needed for reliable aerial-ground matching.
Existing methods address this observation gap either by constructing panoramic ground views from multi-view cameras~\citep{zhu_transgeo_2022, zhang_cross-view_2023, deuser_sample4geo_2023}, or by introducing complementary sensing modalities such as LiDAR to provide near-360$^\circ$ geometric perception and metrically meaningful structural cues~\citep{wang_multi-modal_2025}.

Despite this progress, current aerial-ground place recognition methods still face two key limitations. 
First, existing frameworks rely largely on task-specific visual and point-cloud backbones, leaving foundation-model visual and geometric tokens~\cite{oquab_dinov2_2024, zhang_utonia_2026} underexplored for capturing more representative cross-view cues. 
Second, global descriptors are often produced by static pooling or aggregation mechanisms~\cite{wang_multi-modal_2025, zhu_dcvlaq_2026}, despite the scene-dependent nature of discriminative evidence. 

To address these issues, we propose \textbf{MAG-VLAQ}, a foundation-model-enhanced query aggregation framework for multi-modal aerial-ground cross-view place recognition. 
Inspired by recent advances in ground-view visual place recognition~\cite{ali-bey_boq_2024, izquierdo_salad_2024, zhu_dcvlaq_2026}, MAG-VLAQ introduces pre-trained foundation models for both image and LiDAR point-cloud inputs to obtain more representative local tokens. 
We further investigate how to adapt these foundation models to aerial-ground retrieval, showing that appropriate task-specific adaptation is crucial for aligning visual and geometric tokens across large viewpoint and modality gaps.
At its core, we propose \textbf{ODE-conditioned VLAQ} to construct a more robust global descriptor that is aware of multi-modal fusion. 
Instead of using fixed query centers shared by all observations, we dynamically adapt the learned query centers in a vector of locally aggregated queries (VLAQ) according to the LiDAR-image fused embedding produced by a neural ordinary differential equations (ODE) module, making the aggregation process observation-dependent. 
Finally, we evaluate MAG-VLAQ on KITTI360-AG~\cite{liao_kitti360_2023, wang_multi-modal_2025} and nuScenes-AG~\cite{caesar_nuscenes_2020, wang_multi-modal_2025}, two datasets specifically constructed for aerial-ground place recognition, where our method substantially outperforms existing state-of-the-art (SoTA) methods across both benchmarks.

\begin{figure}[t]
    \centering
    \includegraphics[width=0.8\linewidth]{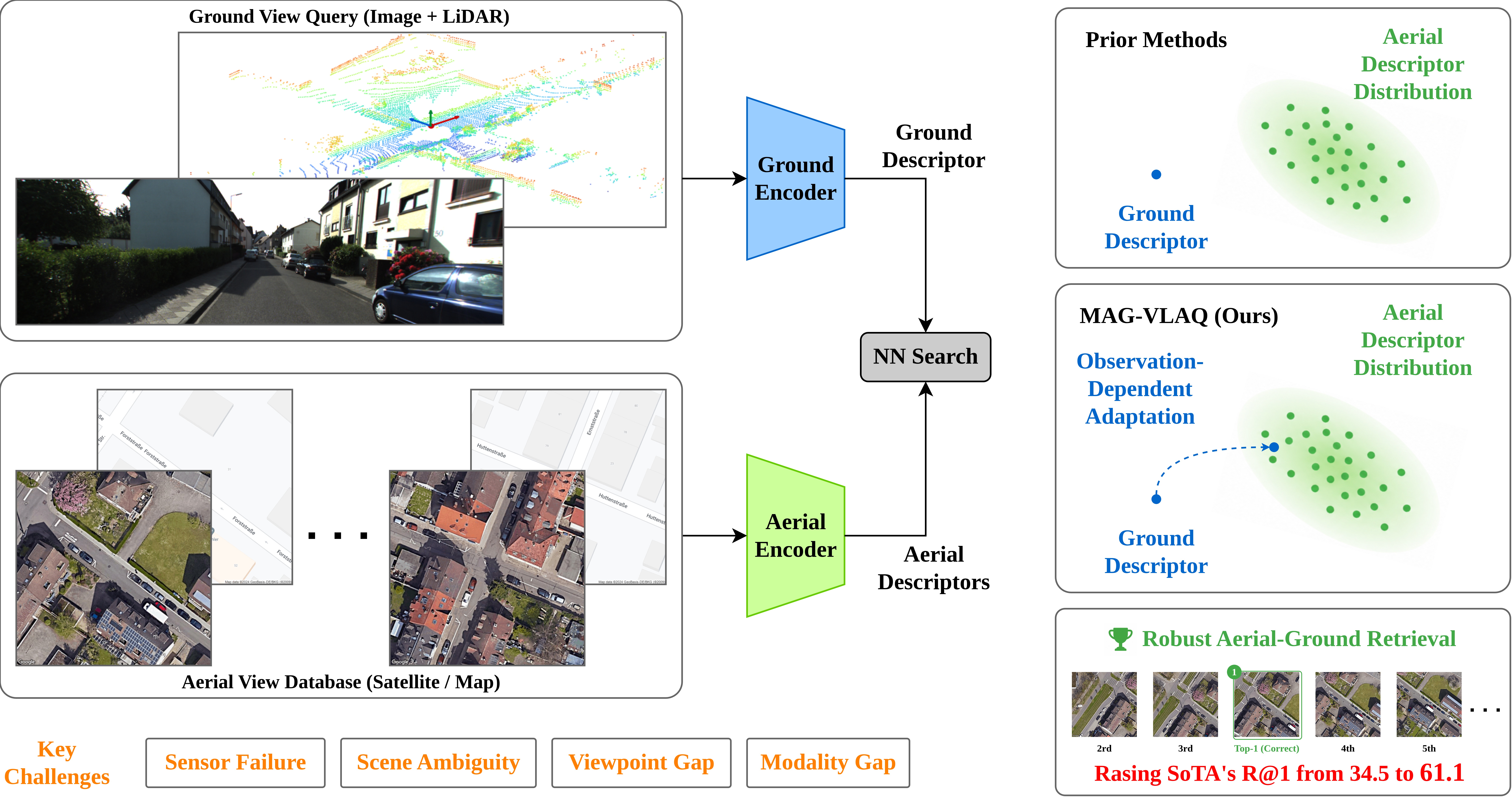}
    \vspace{-1ex}
    \caption{\textbf{We propose MAG-VLAQ for multi-modal aerial-ground place recognition.} Unlike prior methods whose ground descriptors may remain far from the aerial descriptor distribution, MAG-VLAQ introduces observation-dependent adaptive query aggregation to condition descriptor construction on the fused ground representation, bringing ground descriptors closer to the aerial distribution and improving SoTA's Recall@1 from 34.5 to 61.1.}
    \label{fig:teaser}
    \vspace{-2ex}
\end{figure}


\section{Related Work}

\textbf{VFM-based Place Recognition.}
Recent advances in visual foundation models (VFMs)~\cite{oquab_dinov2_2024, radford_learning_2021} have brought new opportunities to place recognition by providing transferable local representations beyond task-specific retrieval backbones. A representative early attempt, AnyLoc~\citep{keetha_anyloc_2023}, shows that off-the-shelf VFM features can generalize well across benchmarks without task-specific training. Following this, recent studies have explored how to better adapt, aggregate, and assign foundation-model tokens for robust retrieval: lightweight adapters, including multi-scale convolutional modules~\citep{lu_cricavpr_2024} and simple multilayer perceptrons~\citep{lu_seamless_2024, lu_selavprplusplus_2025}, improve the discriminability of pre-trained features; query-based methods such as BoQ~\citep{ali-bey_boq_2024} and VLAQ~\citep{zhu_dcvlaq_2026} replace hand-crafted visual words with learnable query prototypes and residual encoding; and assignment-oriented methods, including SALAD~\citep{izquierdo_salad_2024}, SALAD-CM~\citep{izquierdo_close_2024}, and A$^2$GC~\citep{li_a2gc_2025}, improve token-to-cluster matching through optimal transport, clique mining, or geometric constraints. Beyond image-based retrieval, VFMs have also been introduced into LiDAR place recognition by projecting point clouds into image-like representations, as in ImLPR~\citep{jung_imlpr_2025} and Polar Perspectives~\citep{serio_polar_2025}. These methods extend VFM representations to geometry-centric retrieval, but still focus on single-modal descriptors and rely on visual backbones.

\textbf{Multi-modal Place Recognition.}
Multi-modal place recognition improves retrieval robustness by exploiting complementary sensing cues, with cameras providing appearance and semantic information and LiDAR capturing metric geometry and structural layout. Early camera-LiDAR methods, such as MinkLoc++~\citep{komorowski_minkloc_2021}, typically encode each modality separately and fuse their descriptors at a later stage. Subsequent works introduce more explicit fusion mechanisms: PIC-Net~\citep{lu_pic-net_2020} uses attention to fuse image and point-cloud features, AdaFusion~\citep{lai_adafusion_2021} learns adaptive modality weights under different environmental conditions, and LCPR~\citep{zhou_lcpr_2024} combines multi-view RGB images and LiDAR range images through multi-scale attention. More recently, UMF~\citep{garcia-hernandez_umf_2024} unifies local and global vision-LiDAR features with cross-attention and local re-ranking for aliased and low-texture environments. Beyond direct sensor fusion, cross-modal and heterogeneous place recognition further consider retrieval across different sensing modalities or flexible sensor configurations. ModaLink~\citep{xie_modalink_2024} learns modality-consistent descriptors for image-to-point-cloud retrieval, VXP~\citep{li_vxp_2025} explores voxel-cross-pixel alignment for camera-LiDAR recognition, and recent universal multimodal methods incorporate richer inputs such as multi-camera images, LiDAR, semantic masks, text descriptions, or arbitrary sensor combinations~\citep{melekhin_mssplace_2025, xia_uniloc_2024, qi_unimpr_2025}.

\textbf{Cross-view Place Recognition.}
Early aerial-ground cross-view methods mainly follow an image-to-image retrieval paradigm, learning shared ground-aerial embeddings or reducing the geometric gap through orientation cues, polar transformations, spatial-aware aggregation, and feature transport~\citep{workman_wide-area_2015, hu_cvm-net_2018, liu_lending_2019, shi_spatial-aware_2019, shi_optimal_2020}. Subsequent works further improve cross-view matching with local pattern modeling, transformer-based feature learning, geometric layout disentanglement, realistic retrieval benchmarks, and hard-negative sampling~\citep{wang_each_2021, yang_cross-view_2021, zhu_transgeo_2022, zhang_cross-view_2023, zhu_vigor_2021, deuser_sample4geo_2023}. Recent studies have moved toward more realistic and robust cross-view localization settings. FRGeo~\citep{zhang_frgeo_2024} aligns geometric spatial layouts through feature recombination, ConGeo~\citep{mi_congeo_2024} improves robustness to ground-view orientation and field-of-view variations, Panorama-BEV Co-Retrieval~\citep{ye_cross-view_2024} introduces a BEV branch to bridge panorama and satellite views, and semi-supervised or unsupervised paradigms reduce the reliance on precisely labeled ground-satellite pairs~\citep{li_unleashing_2024}. While these methods largely remain image-to-image frameworks, AGPlace~\citep{wang_multi-modal_2025} extends aerial-ground place recognition to a multi-modal setting by integrating ground-view camera and LiDAR observations with a Neural ODE-based fusion framework.

\section{Methodology}
\label{sec:method}

\subsection{Problem Formulation}
\label{sec:problem}
Following standard practice in place recognition~\citep{ali-bey_boq_2024, izquierdo_salad_2024, wang_multi-modal_2025, zhu_dcvlaq_2026}, we formulate multi-modal aerial-ground cross-view place recognition as a ground-to-aerial a retrieval problem.
Given a ground observation as the query, the goal is to retrieve aerial references from a geo-referenced database that correspond to the same physical location.

Let the ground query set $\mathcal{Q}$ and aerial reference database $\mathcal{D}$ be
\begin{equation}    
\mathcal{Q}=\{\bm{q}_i\}_{i=1}^{N_q},\quad
\bm{q}_i=(I_i^g,P_i^g,\mathbf{p}_i^g);
\qquad
\mathcal{D}=\{\bm{r}_j\}_{j=1}^{N_d},\quad
\bm{r}_j=(\{I_{j,m}^a\}_{m=1}^{M},\mathbf{p}_j^a),
\end{equation}
where \(I_i^g \in \{0, \hdots, 255\}^{3 \times w \times h}\) and \(P_i^g \in \mathbb{R}^{3\times n_i}\) denote respectively the \(i^\text{th}\) ground RGB image and the \(i^\text{th}\) LiDAR cloud, composed of \(n_i\) points, \(\{I_{j,m}^a\}_{m=1}^{M} \in \{0, \hdots, 255\}^{3 \times w \times h\times M}\) is the set of \(M\) aerial images in terms of satellite images or road maps, and \(\bm{p}_i^g,\bm{p}_j^a\in\mathbb{R}^2\) are the corresponding geo-locations in east-north coordinates.

Our model extracts one descriptor from each ground and aerial reference, denoted as \(\bm{z}_i^g\) and \(\bm{z}_j^a\), respectively. 
Place similarity is measured by their distance in the shared descriptor space:
\begin{equation}
d_{\mathrm{desc}}(i,j)=\|\bm{z}_i^g-\bm{z}_j^a\|_2.
\end{equation}

Geo-locations are used to mine positive and negative pairs during training, and to evaluate whether a retrieved aerial reference is correct during testing. 
Specifically, an aerial reference is considered positive or correctly retrieved if its geographic distance to the query is smaller than a threshold \(\tau_p\), and negative if it is larger than \(\tau_n\):
\begin{equation}
\mathcal{P}(i)=
\left\{j \mid \|\bm{p}_i^g-\bm{p}_j^a\|_2 < \tau_p \right\},
\quad
\mathcal{N}(i)=
\left\{j \mid \|\bm{p}_i^g-\bm{p}_j^a\|_2 > \tau_n \right\}.
\end{equation}
Note that we deliberately set $\tau_n > \tau_p$ to avoid the ambiguous supervision.

\subsection{Preliminary: VLAQ}
VLAQ is a query-residual aggregator designed to produce robust global descriptors from local tokens, and has demonstrated strong effectiveness in aggregating complementary foundation-model features for place recognition~\citep{zhu_dcvlaq_2026}. 
We briefly review VLAQ to make this paper self-contained.

Given local tokens \(\mathcal{X}=\{\bm{x}_n\}_{n=1}^{N}\) and learnable query prototypes \(\mathcal{C}=\{\bm{c}_s\}_{s=1}^{S}\), VLAQ first computes the assignment weight of each token to each query:
\begin{equation}
e_{n,s}=
\frac{\bm{x}_n^\top\bm{c}_s}{\sqrt{D}},
\qquad
\alpha_{n,s}=
\frac{\exp(e_{n,s})}{\sum_{r=1}^{N}\exp(e_{r,s})}.
\end{equation}
For each query prototype, VLAQ then aggregates token-to-query residuals:
\begin{equation}
\bm{v}_s=
\sum_{n=1}^{N}
\alpha_{n,s}(\bm{x}_n-\bm{c}_s).
\end{equation}
The residual vectors from all queries are normalized, concatenated, and projected to obtain the final global descriptor. 
MAG-VLAQ uses this VLAQ interface for ground and aerial tokens, while further adapting the ground-side query prototypes to each RGB-LiDAR observation, as described in Sec.~\ref{sec::odec-vlaq}.

\begin{figure}[t]
    \centering
    \includegraphics[width=\linewidth]{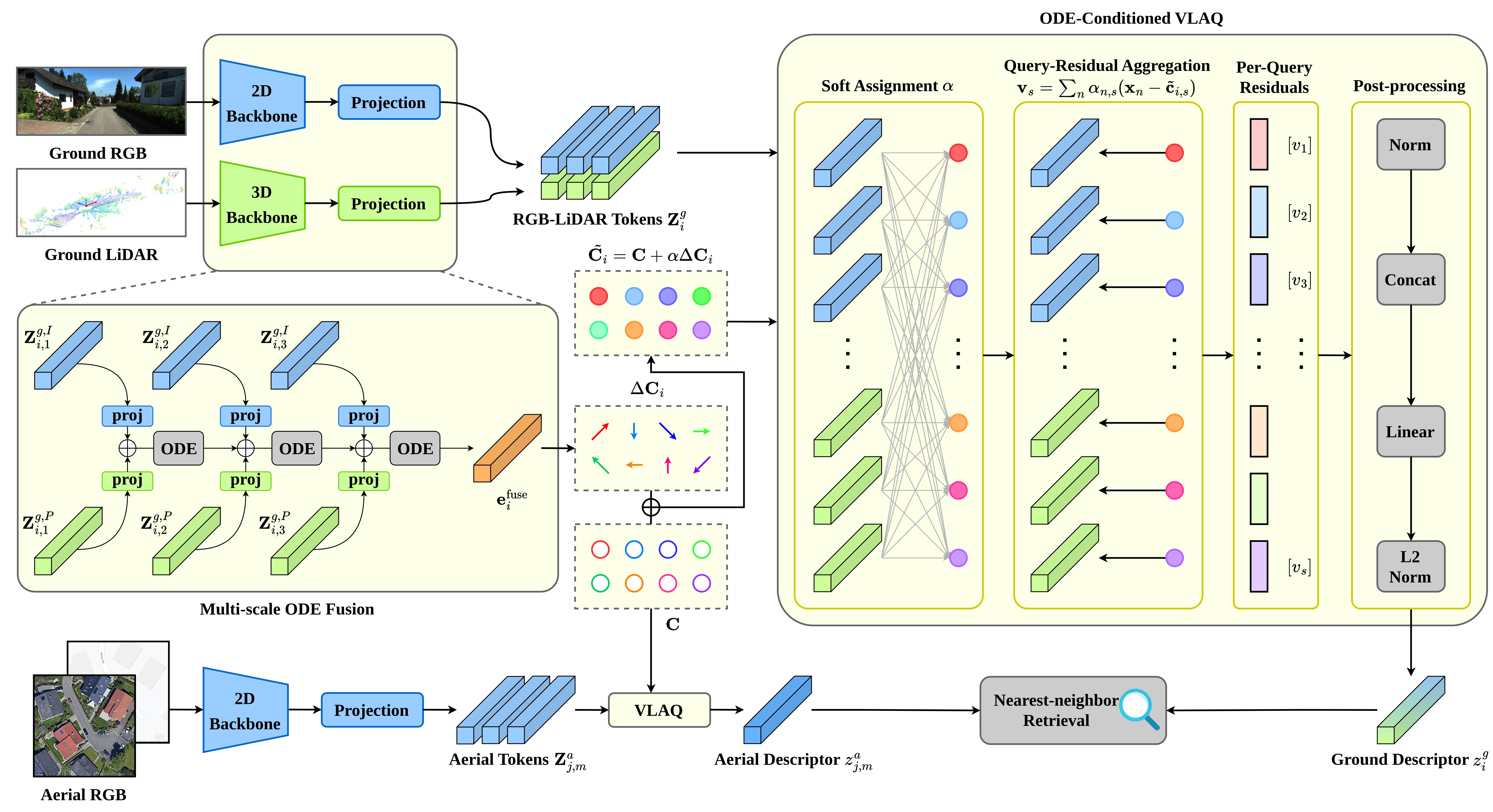}
    \vspace{-4ex}
    \caption{
    \textbf{Overview of MAG-VLAQ.} Ground RGB and LiDAR inputs are encoded with foundation models into local tokens and fused by a multi-scale ODE module to produce a fused feature. This feature conditions VLAQ query centers, enabling observation-dependent query-residual aggregation for the ground descriptor. Aerial images are encoded and aggregated by a shared VLAQ to form database descriptors, followed by nearest-neighbor retrieval for aerial-ground place recognition.
    }
    \label{fig:overview}
    \vspace{-2ex}
\end{figure}

\subsection{Foundation-Model-Enhanced Framework}
\label{sec:framework}

MAG-VLAQ, as shown in Fig.~\ref{fig:overview}, adopts an asymmetric dual-branch framework for aerial-ground cross-view retrieval. 
The ground branch takes an RGB image and a LiDAR point cloud as input, while the aerial branch takes geo-referenced map images as database references. 
The core idea is to extract representative local tokens from each modality using pre-trained foundation models, project them into a shared token space, and then aggregate them into global descriptors.

For the image modality, we use DINOv2~\citep{oquab_dinov2_2024} as the encoder for both ground RGB images and aerial map images. 
For the LiDAR modality, we use Utonia~\citep{zhang_utonia_2026} as the point-cloud encoder to extract geometry-aware local tokens. 
Note that although ground and aerial image encoders share the same DINOv2 architecture, they are adapted independently to account for the large cross-view domain gap. 
Specifically, we fine-tune the last two blocks of each DINOv2 encoder and the last block of the Utonia encoder, while keeping the remaining backbone parameters frozen. 
The extracted local tokens are then mapped to a common aggregation dimension using lightweight MLP projectors:
\begin{equation}
\{\bm{Z}^{g,I}_{i,\ell}\}_{\ell=1}^{L}
=
\Pi_I^g\!\left(\Phi_I^g(I_i^g)\right),
\qquad
\{\bm{Z}^{g,P}_{i,\ell}\}_{\ell=1}^{L}
=
\Pi_P\!\left(\Phi_P(P_i^g)\right),
\qquad
\bm{Z}^{a}_{j,m}
=
\Pi_I^a\!\left(\Phi_I^a(I_{j,m}^a)\right),
\end{equation}
where \(\Phi_I^g\), \(\Phi_I^a\), and \(\Phi_P\) denote the ground image, aerial image, and LiDAR foundation encoders, and \(\Pi_I^g\), \(\Pi_I^a\), and \(\Pi_P\) are modality-specific projectors.

For later descriptor aggregation, we use the last-layer projected tokens. 
On the ground side, the projected RGB and LiDAR tokens are normalized and concatenated along the token dimension:
\begin{equation}
\bm{Z}_i^g
=
\operatorname{LN}(\bm{Z}^{g,I}_{i,L})
\oplus
\operatorname{LN}(\bm{Z}^{g,P}_{i,L}),
\end{equation}
where $\operatorname{LN}$ denotes Layer Normalization and \(\oplus\) denotes token-wise concatenation. 
On the aerial side, each map modality is independently encoded and projected to obtain \(\bm{Z}^{a}_{j,m}\). 
These projected local tokens serve as the inputs to the ODE-Conditioned VLAQ aggregation described in Sec.~\ref{sec::odec-vlaq}.

Moreover, to better exploit the complementary information between ground RGB and LiDAR observations, we adopt the Neural ODE fusion process introduced in AGPlace~\citep{wang_multi-modal_2025}. 
This module operates on multi-scale ground tokens and produces a fused feature that summarizes scene-specific cross-modal information. 
For each scale \(\ell\), the projected RGB and LiDAR tokens are first pooled and transformed by lightweight MLPs:
\begin{equation}
\bm{m}_{i,\ell}
=
h_{\ell}^{I}\!\left(\operatorname{Pool}(\bm{Z}^{g,I}_{i,\ell})\right)
+
h_{\ell}^{P}\!\left(\operatorname{Pool}(\bm{Z}^{g,P}_{i,\ell})\right),
\end{equation}
where \(h_{\ell}^{I}\) and \(h_{\ell}^{P}\) project the image and LiDAR features into the fusion space. 
Starting from the deepest scale, the fusion state is progressively updated through a sequence of Neural ODE blocks:
\begin{equation}
\gamma_i^{\ell}(0)=
\begin{cases}
\bm{m}_{i,L}, & \ell=L,\\
\bm{m}_{i,\ell}+\gamma_i^{\ell+1}(T), & \ell<L,
\end{cases}
\qquad
\frac{d\gamma_i^{\ell}(t)}{dt}
=
F_{\theta_\ell}(\gamma_i^{\ell}(t)).
\end{equation}
After the multi-scale evolution, the final fused feature is obtained as
\begin{equation}
\bm{e}_i^{\mathrm{fuse}}=\gamma_i^{1}(T).
\end{equation}
This fused feature captures scene-specific interactions between visual appearance and LiDAR geometry, which is useful because the discriminative cues for aerial-ground retrieval may vary across scenes. 
In our MAG-VLAQ, \(\bm{e}_i^{\mathrm{fuse}}\) is not used directly as the final descriptor. 
Instead, it provides the conditioning signal to adapt the VLAQ query prototypes, as described in Sec.~\ref{sec::odec-vlaq}.

\subsection{ODE-Conditioned VLAQ}
\label{sec::odec-vlaq}

A key design of MAG-VLAQ is to make query-residual aggregation adaptive to each ground observation. 
To this end, we propose ODE-Conditioned VLAQ, which adapts the ground-side VLAQ~\citep{zhu_dcvlaq_2026} query prototypes according to the RGB-LiDAR fused state. 
A straightforward way to use the fused feature is to append it to the final descriptor as an additional global cue. 
However, such late fusion does not change how local RGB-LiDAR tokens are aggregated. 
MAG-VLAQ instead uses the fused feature to condition the aggregation interface itself, allowing the query prototypes to adapt before residual aggregation is performed.
In this manner, the descriptor is still generated through query-residual aggregation over local tokens, while the query prototypes become observation-dependent.

Specifically, given the observation-specific fused feature \(\bm{e}_i^{\mathrm{fuse}}\) produced by the Neural ODE fusion module, the ground-side VLAQ prototypes are adapted as
\begin{equation}
\tilde{\bm{C}}_i = \bm{C} + \alpha \Delta \bm{C}_i,
\qquad
\Delta \bm{C}_i = \Pi_e(\bm{e}_i^{\mathrm{fuse}}),
\end{equation}
where \(\bm{C}=\{\bm{c}_s\}_{s=1}^{S}\) denotes the shared VLAQ query prototypes, \(\Delta \bm{C}_i\) is the observation-specific query shift predicted from the fused feature by a lightweight MLP \(\Pi_e\), and \(\alpha\) controls the magnitude of query adaptation.

The adapted prototypes \(\tilde{\bm{C}}_i\) and the shared prototypes \({\bm{C}}_i\) are then used to aggregate the concatenated ground RGB-LiDAR tokens \(\bm{Z}_i^g\) and aerial image tokens \(\bm{Z}_{j,m}^a\) respectively with:
\begin{equation}
\bm{z}_i^g
=
\mathcal{A}_{\tilde{\bm{C}}_i}(\bm{Z}_i^g),
\qquad
\bm{z}_{j,m}^{a}
=
\mathcal{A}_{\bm{C}}(\bm{Z}_{j,m}^{a}),
\end{equation}
where \(\mathcal{A}_{\mathcal{C}}(\cdot)\) denotes VLAQ aggregation defined in Eq.~(4)--(5) using the designated query prototypes. 

Following the standard VLAQ formulation, the residual vectors from all prototypes are normalized, concatenated, and projected to obtain the final descriptor:
\begin{equation}
\bm{z}_i^{\Xi}
=
\operatorname{Proj}
\left(
\operatorname{Concat}
\left[
\operatorname{Norm}(\bm{v}_{i,1}^{\Xi}),\ldots,
\operatorname{Norm}(\bm{v}_{i,S}^{\Xi})
\right]
\right), \quad \Xi\in(g, a).
\end{equation}

Overall, this asymmetric design preserves a shared retrieval space while encouraging the ground descriptor to better align with the aerial descriptor distribution. 
By keeping the aerial branch anchored to the shared prototypes \(\bm{C}\), MAG-VLAQ allows aerial descriptors to be pre-computed and indexed offline. 
At the same time, the observation-specific shift \(\Delta \bm{C}_i\) adapts only the ground-side query prototypes, so that RGB-LiDAR fused context can guide ground token aggregation toward the aerial retrieval space.

\subsection{Training Objective}

MAG-VLAQ is trained with ground-to-aerial ranking supervision, coordinate-derived distance consistency, and query-shift regularization. 
For each ground query, we construct triplets \((i,j^+,j^-)\), where \(a_{j^+}\) and \(a_{j^-}\) denote positive and negative aerial references, respectively. 
The triplet ranking loss is defined as
\begin{equation}
\mathcal{L}_{\mathrm{tri}}
=
\frac{1}{|\mathcal{T}|}
\sum_{(i,j^+,j^-)\in\mathcal{T}}
\left[
m+D(i,j^+)-D(i,j^-)
\right]_+,
\end{equation}
where \(D(i,j)\) measures the descriptor distance between the \(i\)-th ground observation and the \(j\)-th aerial reference, \(m\) is the ranking margin, and \([\cdot]_+=\max(\cdot,0)\).

Beyond relative ranking supervision, MAG-VLAQ follows AGPlace~\citep{wang_multi-modal_2025} and adopts a multi-domain distance consistency loss \(\mathcal{L}_{\mathrm{aux}}\) over aerial descriptors, final ground descriptors, and image-only/LiDAR-only ground descriptors. 
This term regularizes descriptor geometry using coordinate-derived spatial relationships, encouraging geographically close observations to be nearby in the embedding space while reducing modality dominance.

To prevent the adapted ground prototypes from drifting excessively from the shared VLAQ prototypes, we further regularize their magnitude with
\begin{equation}
\mathcal{L}_q
=
\frac{1}{B}\sum_i\|\Delta\bm{C}_i\|_F^2.
\end{equation}
 
Overall, the complete training objective is given by
\begin{equation}
\mathcal{L}
=
\lambda_{\mathrm{tri}}\mathcal{L}_{\mathrm{tri}}
+
\lambda_{\mathrm{aux}}\mathcal{L}_{\mathrm{aux}}
+
\lambda_q\mathcal{L}_{q}.
\end{equation}

\section{Experiments}
\label{sec:experiments}

\subsection{Datasets, Baselines, and Implementation Details}
\label{sec:exp_details}
\textbf{Datasets.} To evaluate MAG-VLAQ, we follow AGPlace~\citep{wang_multi-modal_2025} and use two multi-modal aerial-ground place recognition benchmarks: KITTI360-AG and nuScenes-AG. 
\textbf{KITTI360-AG} is built from the urban driving sequences of KITTI360~\citep{liao_kitti360_2023}. 
Each ground observation contains a LiDAR point cloud, an image from a forward-facing RGB camera, and the corresponding GNSS coordinate. 
The aerial references include satellite images and road maps retrieved according to the GNSS locations of the ground frames. Both \textit{ground-to-satellite} and \textit{ground-to-map} retrieval are evaluated.
\textbf{nuScenes-AG} is constructed from nuScenes~\citep{caesar_nuscenes_2020}, another large-scale urban driving dataset. 
Each ground observation provides a LiDAR point cloud, GNSS coordinate, and images from six surrounding-view cameras. 
Its aerial database contains satellite images only, generated in the same manner as KITTI360-AG using the GNSS coordinates of the ground frames.
We evaluate \textit{ground-to-satellite} retrieval and further conduct \textit{sensor-failure} experiments.
For both datasets, we follow the train/test splits of AGPlace~\citep{wang_multi-modal_2025}.

\textbf{Baselines.} 
For ground-to-satellite and ground-to-map evaluation on KITTI360-AG, we compare MAG-VLAQ with a broad set of place recognition methods, including image-based methods~\citep{ali-bey_gsvcities_2022, ali-bey_mixvpr_2023, keetha_anyloc_2023, izquierdo_salad_2024, ali-bey_boq_2024, zhu_dcvlaq_2026}, point-cloud-based methods~\citep{komorowski_minkloc3dv2_2022, luo_bevplace_2023}, and multi-modal methods~\citep{komorowski_minkloc_2021, lai_adafusion_2021, melekhin_mssplace_2025, zhou_lcpr_2024, garcia-hernandez_umf_2024}. 
We also include recent methods specifically designed for cross-view or cross-modal place recognition, including TransGeo~\citep{zhu_transgeo_2022}, Sample4Geo~\citep{deuser_sample4geo_2023}, ArcGeo~\citep{shugaev_arcgeo_2024}, ConGeo~\citep{mi_congeo_2024}, VXP~\citep{li_vxp_2025}, Lip-Loc~\citep{shubodh_lip-loc_2024}, and AGPlace~\citep{wang_multi-modal_2025}. 
For ground-to-satellite and sensor-failure experiments on nuScenes-AG, we mainly compare with multi-modal methods~\citep{komorowski_minkloc_2021, lai_adafusion_2021, melekhin_mssplace_2025, zhou_lcpr_2024, garcia-hernandez_umf_2024, wang_multi-modal_2025}.

\textbf{Implementation Details.} MAG-VLAQ uses DINOv2~\citep{oquab_dinov2_2024} as the image foundation encoder and Utonia~\citep{zhang_utonia_2026} as the LiDAR token extractor. 
Ground and aerial images are resized to \(224\times224\), and all output descriptors are L2-normalized to 512. 
For ODE-Conditioned VLAQ, we follow the standard setup of BoQ~\citep{ali-bey_boq_2024} and VLAQ~\citep{zhu_dcvlaq_2026} and use 64 learnable queries. 
During training, the thresholds for positive and negative sample mining are set to \(\tau_p=10\,\mathrm{m}\) and \(\tau_n=25\,\mathrm{m}\), respectively. 
We train MAG-VLAQ for 40 epochs using Adam with a learning rate of \(1\times10^{-6}\). 
All experiments are conducted on 4 NVIDIA GeForce RTX 4090 GPUs.
For evaluation, aerial references within 25 m of each ground query are treated as correct retrievals, and Recall@K with \(K=\{1,5,10\}\) is reported for all methods under the same dataset splits and retrieval protocol.

\subsection{Main Results}
\label{sec:main_results}
\textbf{KITTI360-AG (Satellite and Road Map).}
Tab.~\ref{tab:kitti360_ag_results} reports quantitative comparisons on KITTI360-AG with satellite images and road maps as aerial references. 
MAG-VLAQ achieves the best performance under both retrieval settings. 
Compared with the strongest previous multi-modal baseline AGPlace, MAG-VLAQ improves satellite R@1 from 32.0\% to 61.1\%, and road map R@1 from 28.2\% to 42.6\%, with consistent gains at higher recall levels. 

The improvement is larger on satellite retrieval, since satellite images preserve richer appearance, texture, and spatial layout cues that can be better exploited by foundation-model tokens. 
Road maps contain less detail and mainly contain structural information, making cross-view matching more challenging. 
Nevertheless, the consistent gains on road map retrieval suggest that MAG-VLAQ does not simply rely on visual appearance, but also benefits from ODE-conditioned VLAQ aggregation that captures geometry-aware and structure-aware correspondences across aerial reference modalities.

\begin{table}[t]
\centering
\caption{\textbf{Quantitative results on the KITTI360-AG dataset} using satellite or road map aerial references. 
"C" denotes the model is designed for cross-view/cross-modal place recognition.
The best results are shown in \textbf{bold} and the second best are \underline{underlined}.
}
\begin{tabular}{l|c|c|c}
\toprule
\textbf{Model} 
& \textbf{Type} & \textbf{Satellite R@1/5/10} & \textbf{Road Map R@1/5/10} \\
\midrule
ConvAP~\citep{ali-bey_gsvcities_2022} 
& 2D   & 25.2 / 40.8 / 48.6 & 20.3 / 32.0 / 39.7 \\
MixVPR~\citep{ali-bey_mixvpr_2023} 
& 2D   & 19.4 / 30.5 / 37.3 & 19.3 / 32.0 / 39.1 \\
AnyLoc~\citep{keetha_anyloc_2023} 
& 2D   & 4.2 / 8.7 / 13.0   & 5.4 / 5.5 / 5.9 \\
SALAD~\citep{izquierdo_salad_2024} 
& 2D   & 26.1 / 34.3 / 39.3 & 17.0 / 27.4 / 33.5 \\
BoQ~\citep{ali-bey_boq_2024} 
& 2D   & 26.9 / 38.9 / 45.1 & 16.4 / 24.9 / 31.0 \\
DC-VLAQ~\citep{zhu_dcvlaq_2026} 
& 2D   & 27.7 / 41.2 / 48.3 & 20.1 / 30.8 / 37.5 \\
TransGeo~\citep{zhu_transgeo_2022} 
& 2D+C & 22.9 / 32.9 / 39.2 & 22.4 / 31.9 / 36.9 \\
Sample4Geo~\citep{deuser_sample4geo_2023} 
& 2D+C & 27.0 / 41.7 / 49.0 & 22.7 / 41.0 / 47.6 \\
ArcGeo~\citep{shugaev_arcgeo_2024} 
& 2D+C & 28.2 / 41.3 / 48.5 & 24.4 / 39.6 / 46.5 \\
ConGeo~\citep{mi_congeo_2024} 
& 2D+C & \underline{34.5} / 40.0 / 44.7 & \underline{29.3} / 36.9 / 41.8 \\
\midrule
MinkLoc3DV2~\citep{komorowski_minkloc3dv2_2022} 
& 3D   & 26.5 / 39.1 / 46.0 & 23.3 / 36.0 / 43.8 \\
BEVPlace~\citep{luo_bevplace_2023} 
& 3D   & 22.3 / 33.9 / 40.8 & 23.6 / 35.5 / 41.4 \\
VXP~\citep{li_vxp_2025} 
& 3D+C & 20.7 / 34.5 / 41.0 & 22.4 / 34.9 / 42.0 \\
Lip-Loc~\citep{shubodh_lip-loc_2024} 
& 3D+C & 29.9 / 42.2 / 49.0 & 24.5 / 35.6 / 42.4 \\
\midrule
MinkLoc++~\citep{komorowski_minkloc_2021} 
& MM   & 28.9 / 39.3 / 44.9 & 26.5 / 40.8 / 48.8 \\
AdaFusion~\citep{lai_adafusion_2021} 
& MM   & 26.5 / 39.6 / 47.8 & 27.2 / 41.6 / 49.3 \\
MSSPlace~\citep{melekhin_mssplace_2025} 
& MM   & 25.5 / 37.4 / 45.0 & 26.7 / 41.2 / 49.0 \\
LCPR~\citep{zhou_lcpr_2024}
& MM   & 27.7 / 44.4 / 50.6 & 24.5 / 39.0 / 47.7 \\
UMF~\citep{garcia-hernandez_umf_2024}
& MM   & 27.1 / 42.6 / 49.2 & 25.6 / 40.4 / 49.7 \\
AGPlace~\citep{wang_multi-modal_2025}
& MM+C & 32.0 / \underline{47.6} / \underline{54.9} & 28.2 / \underline{43.3} / \underline{52.0} \\
\midrule                                   
\textbf{MAG-VLAQ (ours)} 
& MM+C & \textbf{61.1 / 77.8 / 83.0} & \textbf{42.6 / 63.1 / 72.7} \\
\bottomrule
\end{tabular}
\label{tab:kitti360_ag_results}
\vspace{-1ex}
\end{table}

\begin{table*}[t]
\centering
\caption{\textbf{Quantitative results on the nuScenes-AG dataset} using satellite image database. ``fail'' denotes dropping the modality input during testing. All models are trained with both modalities.
The best results are shown in \textbf{bold} and the second best are \underline{underlined}.}
\vspace{1ex}
\setlength{\tabcolsep}{7pt}
\begin{tabular}{@{}l|cc|cc|cc@{}}
\toprule
\multirow{2}{*}{\textbf{Model}}
& \multicolumn{2}{c|}{\textbf{Cams. + LiDAR}}
& \multicolumn{2}{c|}{\textbf{Cams. Fail}}
& \multicolumn{2}{c}{\textbf{LiDAR Fail}} \\
& \textbf{R@1} & \textbf{R@5} & \textbf{R@1} & \textbf{R@5} & \textbf{R@1} & \textbf{R@5} \\
\midrule
MinkLoc++~\citep{komorowski_minkloc_2021} & 70.4 & 82.1 & 16.9 & 27.3 & 2.6 & 7.3 \\
AdaFusion~\citep{lai_adafusion_2021}      & 71.9 & 82.3 & 9.0  & 17.8 & 3.5 & 6.7 \\
MSSPlace~\citep{melekhin_mssplace_2025}   & 71.3 & 82.2 & 17.4 & 29.8 & 5.0 & 8.7 \\
LCPR~\citep{zhou_lcpr_2024}               & 57.7 & 74.2 & 3.2  & 6.2  & 8.6 & 17.1 \\
UMF~\citep{garcia-hernandez_umf_2024}         & 69.4 & 82.5 & 2.7  & 7.3  & 0.8 & 2.9 \\
AGPlace~\citep{wang_multi-modal_2025}     & \underline{75.6} & \underline{87.2} & \underline{22.8} & \underline{33.1} & \underline{12.9}  & \underline{23.0} \\
\midrule                                                
\textbf{MAG-VLAQ (ours)}                          & \textbf{82.6} & \textbf{90.1} & \textbf{48.7} & \textbf{77.2} & \textbf{30.3} & \textbf{50.8} \\
\bottomrule
\end{tabular}
\label{tab:nuscenes_ag_fail}
\vspace{-1ex}
\end{table*}

\textbf{nuScenes-AG (Sensor Failure).}
Tab.~\ref{tab:nuscenes_ag_fail} reports quantitative comparisons on nuScenes-AG. 
MAG-VLAQ achieves the best performance under all sensor settings. 
Especially with sensor failures, MAG-VLAQ improves R@1/R@5 from 22.8\%/33.1\% to 48.7\%/77.2\% under camera failure, and from 12.9\%/23.0\% to 30.3\%/50.8\% under LiDAR failure. 

The strong robustness under sensor failures highlights the benefit of the proposed ODE-conditioned VLAQ. 
When cameras fail, MAG-VLAQ can still aggregate LiDAR-derived structural cues, such as road topology and building boundaries, into a discriminative global descriptor. 
When LiDAR fails, it can better organize foundation-model image tokens to bridge the larger viewpoint and domain gap.

\begin{figure}[t]
\centering
\includegraphics[width=\textwidth]{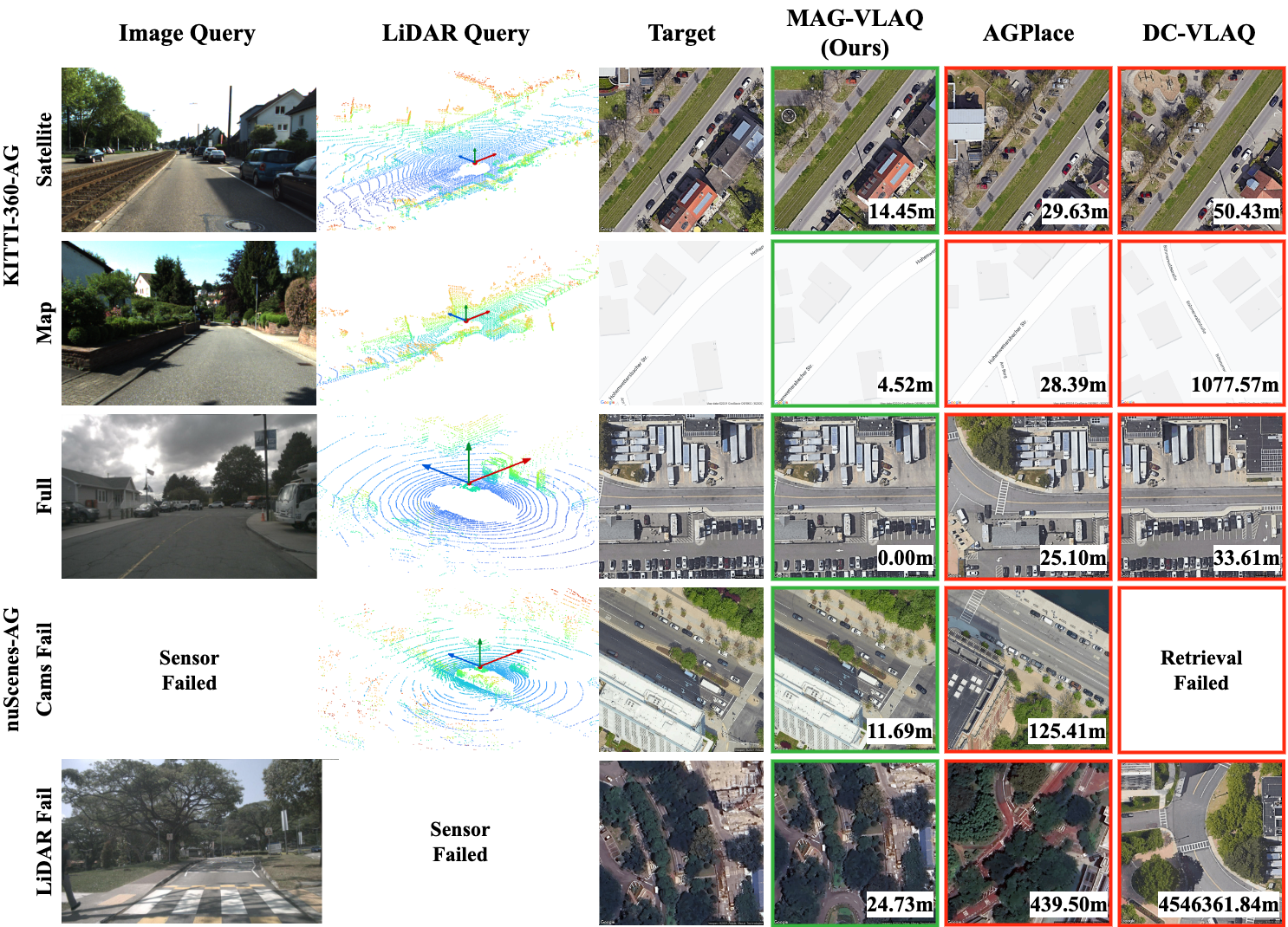}
\caption{
\textbf{Qualitative results of top-1 retrievals under all five experimental settings.} For each query, we show the ground-view image, LiDAR point cloud, target aerial reference, and the top-1 retrieval results of MAG-VLAQ, AGPlace, and DC-VLAQ. \textcolor{ForestGreen}{Green} boxes indicate correct retrievals, while \textcolor{red}{red} boxes indicate incorrect or failed retrievals. The number in the bottom-right corner of each retrieval result denotes the physical distance between the retrieved location and the target location.
}
\label{fig:qualitative_comparison}
\end{figure}

Fig.~\ref{fig:qualitative_comparison} further provides qualitative examples across KITTI360-AG, nuScenes-AG, and sensor-failure settings. The results further demonstrate the robustness of MAG-VLAQ across different reference types and incomplete sensor inputs.

\subsection{Ablation Studies}
\label{sec:ablation}
Tab.~\ref{tab:ablation} reports ablations on KITTI360-AG with the satellite database, examining foundation-model adaptation and global descriptor aggregation.

\begin{table}[t]
\centering
\caption{\textbf{Ablation studies} on foundation-model adaptation and global descriptor aggregation.
The best results are shown in \textbf{bold} and the second best are \underline{underlined}.
}
\resizebox{0.85\linewidth}{!}{%
\begin{tabular}{l|ccc|rrr}
\toprule
\multirow{2}{*}{\textbf{Variant}} 
& \textbf{Image Enc.}
& \textbf{LiDAR Enc.}
& \textbf{Global}
& \multicolumn{3}{c}{\textbf{KITTI360-AG (Sat.)}} \\
& \textbf{Trainable} & \textbf{Trainable} & \textbf{Descriptor} & \textbf{R@1} & \textbf{R@5} & \textbf{R@10} \\
\midrule
Default 
& \textit{Last 2 Blocks} & \textit{Last 1 Block} & \textit{ODE-C VLAQ} & \textbf{61.1} & \textbf{77.8} & \textbf{83.0} \\
\midrule
\multirow{3}{*}{\makecell[l]{Adapt.\\Var.}} 
& \textit{Frozen} & \textit{Frozen} &  & 51.2 & 68.6 & 76.8 \\
& \textit{LoRA} & \textit{LoRA} &  & 53.9 & 72.8 & 79.8 \\
& \textit{MultiConv} & \textit{MLP} &  & 11.4 & 23.3 & 28.9 \\
\midrule
\multirow{3}{*}{\makecell[l]{Global\\Var.}} 
&  &  & \textit{Pooling-based} & 48.4 & 65.6 & 72.2 \\
&  &  & \textit{BoQ} & 52.6 & 69.8 & 76.4 \\
&  &  & \textit{VLAQ} & \underline{56.7} & \underline{74.7} & \underline{81.2} \\
\bottomrule
\end{tabular}
}
\label{tab:ablation}
\vspace{-2ex}
\end{table}

\textbf{Foundation-Model Adaptation.}
We first study how to adapt the image and LiDAR foundation encoders. 
Using frozen backbones gives 51.2/68.6/76.8 R@1/R@5/R@10, while LoRA improves the result to 53.9/72.8/79.8, showing that limited backbone adaptation is helpful for aerial-ground retrieval. 
However, fine-tuning the last two image blocks and the last LiDAR block further increases the performance to 61.1/77.8/83.0, indicating that high-level visual and geometric tokens need more direct task-specific adaptation for the asymmetric ground-to-aerial matching objective. 
In contrast, the MultiConv/MLP adapter performs poorly, suggesting that generic 2D PEFT modules are not directly suitable for this multi-modal cross-view setting, especially when LiDAR tokens depend strongly on geometric structure and positional encoding.

\textbf{Global Descriptor Aggregation.}
We then compare different global descriptor while keeping the same backbone adaptation and local token projection. 
Pooling-based aggregation obtains 48.4/65.6/72.2 R@1/R@5/R@10, which is much lower than static VLAQ at 56.7/74.7/81.2. 
This shows that simple pooling loses important local evidence distributed across visual and geometric tokens. 
By aggregating residuals with respect to learned query prototypes, VLAQ preserves richer token-level structure and produces a more discriminative aerial-ground descriptor. 
ODE-conditioned VLAQ further improves static VLAQ to 61.1/77.8/83.0, confirming that the ODE-fused RGB-LiDAR state provides useful scene-level guidance for query adaptation. 
This supports the central design of MAG-VLAQ: multi-modal fusion should not only produce a fused feature, but also guide how local tokens are assigned and aggregated into the final global descriptor.

\section{Conclusion}
\label{sec:conclusions}
In this work, we propose \textbf{MAG-VLAQ}, a foundation-model-enhanced query aggregation framework for multi-modal aerial-ground cross-view place recognition. 
MAG-VLAQ adapts visual and LiDAR foundation models to extract task-specific local tokens, and introduces ODE-conditioned VLAQ to generate fusion-aware global descriptors through dynamic query-residual aggregation. 
Experiments on KITTI360-AG and nuScenes-AG show that MAG-VLAQ consistently outperforms existing 2D, 3D, multi-modal, and cross-view baselines across different aerial references and sensor-availability settings. 
Ablation studies further validate the importance of foundation-model adaptation and ODE-conditioned query aggregation for robust aerial-ground retrieval. 
Future work will explore cross-view-aware PEFT strategies to adapt visual and geometric foundation models more efficiently while preserving spatial structure and improving aerial-ground alignment.

\bibliographystyle{abbrvnat}
\bibliography{references}

\newpage
\appendix
\section{Attention Visualization}

\begin{figure}[h]
    \centering
    \includegraphics[width=\linewidth]{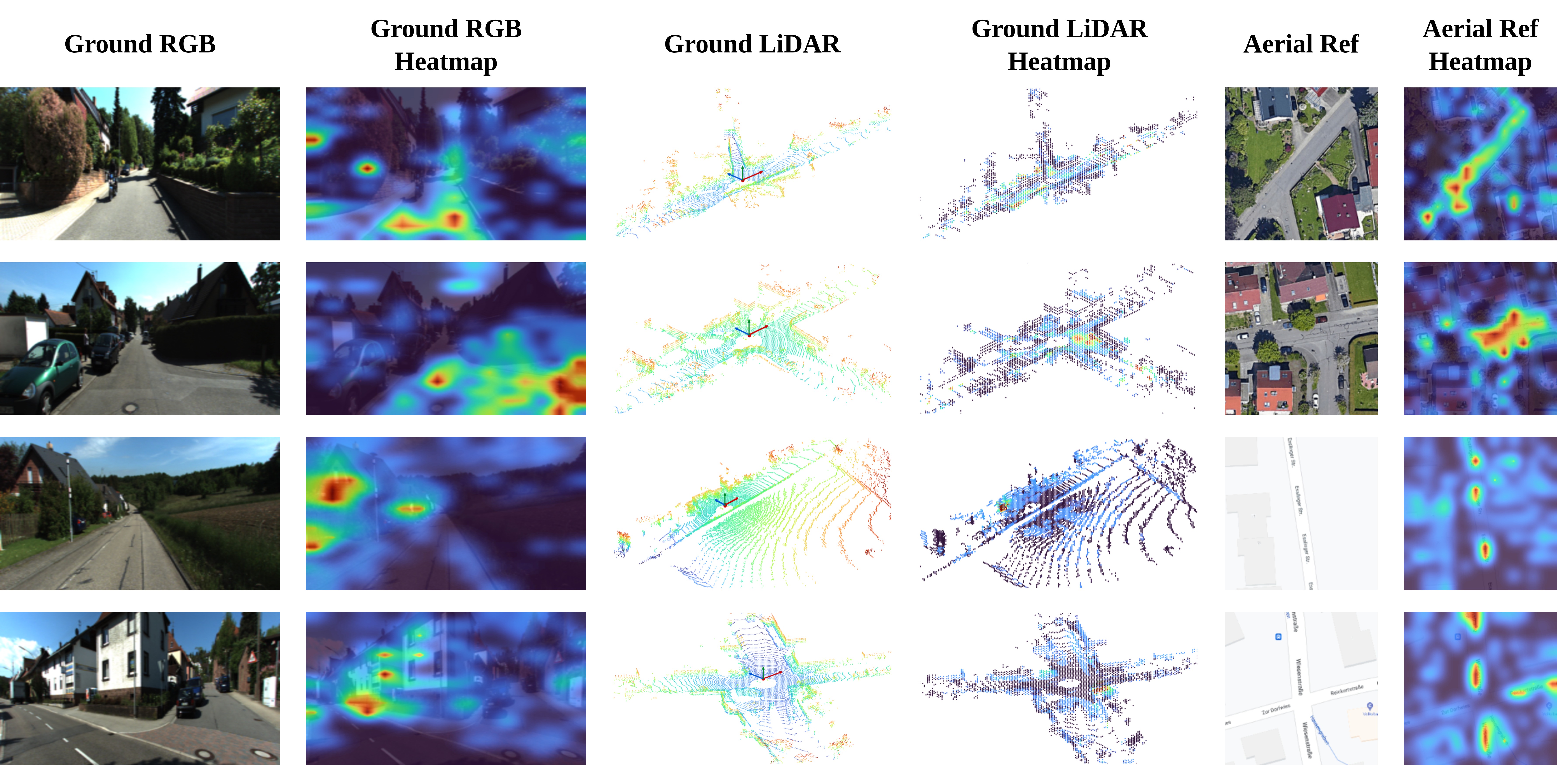}
    \caption{Attention Visualization}
    \label{fig:attn_vis}
\end{figure}

Fig.~\ref{fig:attn_vis} provides qualitative attention visualizations for the ground RGB image, ground LiDAR point cloud, and aerial reference. Warmer regions indicate tokens that receive higher weights during descriptor aggregation. Across different scenes, MAG-VLAQ consistently highlights spatially stable and cross-view transferable cues, such as road surfaces, road boundaries, intersections, building facades, and structural edges. In the LiDAR branch, the responses are mainly concentrated on geometric layouts, including road topology, building boundaries, and building structures, while the aerial heatmaps emphasize corresponding top-down patterns such as road centerlines, junctions, and block-level boundaries. These patterns suggest that the proposed ODE-conditioned VLAQ does not rely only on local appearance, but uses the RGB-LiDAR fused state to guide query--residual aggregation toward scene-specific structural evidence that is more consistent across ground and aerial views.

\section{Limitation Discussion}
\label{app:limitation}
Although MAG-VLAQ achieves strong performance on KITTI360-AG and nuScenes-AG, several limitations remain. First, our evaluation is conducted on urban driving benchmarks, and the generalization ability of MAG-VLAQ to broader geographic regions, rural scenes, highways, adverse weather, nighttime conditions, and different aerial map sources has not been fully verified. Second, the method assumes that ground RGB, LiDAR observations, and geo-referenced aerial references are available and reasonably aligned. In real-world deployment, sensor degradation, sparse LiDAR scans, camera exposure failures, GNSS noise, outdated aerial imagery, or inaccurate map crops may reduce retrieval reliability. Third, MAG-VLAQ uses foundation encoders for both visual and LiDAR inputs and fine-tunes high-level backbone blocks, which introduces additional memory and computational cost compared with lightweight retrieval models. Finally, our task is formulated as place retrieval under a distance threshold, rather than precise metric localization. Therefore, MAG-VLAQ should be viewed as a robust retrieval component that may need to be combined with temporal filtering, geometric verification, or SLAM-based pose estimation for safety-critical localization.

\section{Broader Impact}

MAG-VLAQ has potential positive impacts for robotics, autonomous navigation, map-assisted localization, and long-term place recognition. By improving aerial-ground retrieval with complementary RGB and LiDAR cues, the method may help robots and vehicles localize more robustly in environments where GNSS is unreliable or one onboard sensor is partially unavailable. The sensor-failure results further suggest that multi-modal retrieval systems can provide useful redundancy for practical navigation and mapping applications.

At the same time, stronger cross-view place recognition can also introduce privacy and misuse risks. In particular, aerial-ground localization technology could be used for unwanted location inference, large-scale monitoring, or tracking of vehicles and individuals when combined with other sensing systems. The model may also inherit geographic and environmental biases from the datasets on which it is trained, leading to uneven performance across regions with different road layouts, map quality, or sensing conditions. Moreover, incorrect retrievals could cause downstream localization errors if the system is used without additional verification. Responsible deployment should therefore include privacy-preserving data handling, access control for sensitive applications, evaluation across diverse regions and conditions, and integration with independent localization or safety-checking modules rather than relying on retrieval alone.


\end{document}